\pdfoutput=1

\documentclass[11pt]{article}

\usepackage[]{EACL2023}

\usepackage{times}
\usepackage{latexsym}

\usepackage[T1]{fontenc}

\usepackage[utf8]{inputenc}

\usepackage{microtype}

\usepackage{inconsolata}

\title{Predicting Machine Translation Performance \\
on Low-Resource Languages: 
The Role of Domain Similarity}

\author{Eric Khiu$^*$, Hasti Toossi$^\dagger$, David Anugraha$^\dagger$, Jinyu Liu$^\dagger$, Jiaxu Li$^\dagger$, \\ \textbf{Juan Armando Parra Flores}$^\P$, \textbf{Leandro Arcos Roman}$^\S$, \\
\textbf{A. Seza Doğruöz}$^\#$, \textbf{En-Shiun Annie Lee}$^{\dagger,\ddagger}$\\
$^*$ University of Michigan, USA 
$^\dagger$ University of Toronto, Canada \\
$^\P$ Centro de Investigación en Matemáticas, Mexico
$^\S$ Amherst College, USA \\
$^\#$ LT3, ID-Lab, Universiteit Gent, Belgium
$^\ddagger$ Ontario Tech University, Canada
}

\usepackage{preambles}
\usepackage{tabularx}
\usepackage{ragged2e}
\usepackage{float}
\usepackage{multirow}
\usepackage{siunitx}
\usepackage{longtable}
\usepackage{caption}
\usepackage{subcaption}
\usepackage{xcolor}
\usepackage{arydshln}
\begin{document}
\maketitle
\begin{abstract}
    Fine-tuning and testing a multilingual large language model is expensive and challenging for low-resource languages (LRLs). While previous studies have predicted the performance of natural language processing (NLP) tasks using machine learning methods, they primarily focus on high-resource languages, overlooking LRLs and shifts across domains. Focusing on LRLs, we investigate three factors: the size of the fine-tuning corpus, the domain similarity between fine-tuning and testing corpora, and the language similarity between source and target languages. We employ classical regression models to assess how these factors impact the model's performance. Our results indicate that domain similarity has the most critical impact on predicting the performance of Machine Translation models.   
    
\end{abstract}

\section{Introduction}
Fine-tuning large language models for natural language processing (NLP) tasks across varying languages, tasks, and domains is a resource-intensive and environmentally harmful process. \citep{xia-etal-2020-predicting}.
This challenge is especially magnified for low-resource languages (LRLs). 
However, knowing how well a language model performs on a particular language can be useful information, such as improving the accuracy of quality estimation (QE) models \citep{zouhar-etal-2023-poor}. 
Therefore, there is a need to estimate the performance of these models for LRLs without conducting time-consuming and computationally expensive model pre-training and fine-tuning.


Existing approaches for predicting the performance of models for NLP tasks have shown promise using linear regression and gradient-boosting trees \cite{birch-etal-2008-predicting,xia-etal-2020-predicting, srinivasan2021predicting, ye-etal-2021-towards}.
These studies have considered data size, typological features, and language similarity as factors contributing to the model performance. However, most of these studies are conducted for high-resource languages (HRLs)
(e.g., Romance and Germanic families) thus limiting their applicability to LRLs. 
Furthermore, performance drops in NLP tasks have been observed due to domain shift\citep{elsahar-galle-2019-annotate}. However, this factor is not explicitly considered in the existing works that predict the performance of language models. 

Based on the aforementioned limitations in the literature, we considered three factors for the Machine Translation (MT) performance prediction for LRLs using classical regression models. These factors are the size of the fine-tuning corpus, the domain similarity between fine-tuning and testing corpora, and the language similarity between source and target languages.

Then, we tested the statistical reliability of these regression models and evaluated them based on their prediction accuracy. We selected those with relatively high accuracy for each factor and explored how data partitioning (described in \S~\ref{sec:model-and-data}) affects the quality of fit using these preferred models. Additionally, we analyzed the importance of the factors by ranking them based on their correlation with the MT performance, their weights in multifactor regression models, and their importance in multifactor models using the Random Forest Regressor.

Our contributions are as follows: 1) we developed a statistically rigorous method for performance prediction that can be repeated on any combination of LRLs, NLP tasks, and LLMs; 2) we specifically evaluated the impact of various factors on the performance of MT models; 3) we provided domain-specific and language-specific interpretations based on the performance of the regression models.

\section{Model and Data}
\label{sec:model-and-data}
Our data is collected from experiments of a prior study \citep{nayak2023leveraging} on fine-tuning and testing different corpora and target languages using the multilingual large language model mBART (Table~\ref{raw-expr-records}). Each experiment consists of performance measured by spBLEU, with the source language (always English (\textsc{en})), the target language, $l$, the fine-tuning corpus, $t$ and its size, $s$, and the testing corpus, $\tau$. 

\begin{table*}[htp!]
\centering
\resizebox{\textwidth}{!}{%
\begin{tabular}{l|l|lll|lll|lll|lll|lll}
\hline
\multirow{3}{0.5in}{Fine-Tuning Corpus} & \multirow{3}{*}{Size} & \multicolumn{15}{c}{Target Language and Testing Corpus} \\
\cline{3-17}
 &  & \multicolumn{3}{|c}{Kannada (\textsc{ka})} & \multicolumn{3}{|c}{Gujarati (\textsc{gu})} & \multicolumn{3}{|c}{Hindi (\textsc{hi})} & \multicolumn{3}{|c}{Sinhala (\textsc{si})} & \multicolumn{3}{|c}{Tamil (\textsc{ta})} \\ 

 & & \textsc{Flores} & Bible & PMI & \textsc{Flores} & Bible & PMI & \textsc{Flores} & Bible & PMI & \textsc{Flores*} & Bible$^\dagger$ & Gov & \textsc{Flores} & Bible & Gov \\
\hline
\multirow{4}{*}{Gov/PMI} & 1k & 2.2 & 0.3 & 12.0 & 7.8 & 2.3 & 22.6 & 6.6 & 1.0 & 19.7  & 3.8 & 0.2 & 21.7 & 2.6 & 0.3 & 19.7 \\
& 10k & 11.8 & 1.5 & 30.7  & 16.6 & 4.0 & 34.2 & 14.5 & 3.0 & 32.4  & 9.2 & 0.9 & 41.7 & 7.1 & 0.8 & 34.8 \\
& 25k & 14.2 & 1.7 & 34.3 & 19.9 & 4.8 & 37.9 & 17.0 & 3.5 & 35.5 & 11.3 & 1.2 & 47.0 & 9.0 & 1.3 & 38.2 \\
& 50k & NA & NA & NA & NA & NA & NA & 19.0 & 3.4 & 36.7 & 12.3 & 1.5 & 49.5 & 11.3 & 1.6 & 40.8 \\
\hline
\multirow{3}{*}{Bible} & 1k  & 0.5 & 12.3 & 0.3 &  2.2 & 12.9 & 1.8 & 1.5 & 18.6 & 1.0 & 0.8 & 21.6 & 0.4 & 0.8 & 16.3 & 0.3 \\
& 10k & 1.8 & 24.0 & 0.8 & 4.1 & 23.9 & 2.6 & 2.5 & 28.1 & 1.8 & 1.7 & 34.2 & 0.8 & 1.6 & 26.9 & 0.7 \\
& 25k & 2.2 & 28.1 & 1.0 & 4.2 & 28.5 & 2.9 & 2.8 & 32.3 & 1.8 & 1.9 & 38.5 & 0.9 & 2.0 & 31.4 & 0.8 \\
\hline 
\end{tabular}
}
\caption{\label{raw-expr-records} MT Performance in spBLEU by fine-tuning mBART on different combinations of fine-tuning corpus, size of fine-tuning corpus, target language, and testing corpus. \\
\footnotesize{* We used $\textsc{Flores-v1}$ instead of $\textsc{Flroes-101}$ for $\textsc{si}$ due to availability. \\
$^\dagger$ The bible corpus for $\textsc{si}$ is scrapped from a different website due to availability.}}
\end{table*}

\paragraph{Language Model and Evaluation Metric} mBART is a pre-trained multilingual sequence-to-sequence model that is built based on the encoder-decoder Transformer architecture \citep{vaswani2017attention}. \newcite{lee-etal-2022-pre} has shown that mBART outperforms mT5, another multilingual large language model, especially on LRLs. \newcite{lee-etal-2022-pre} also suggested the use of spBLEU as the evaluation metric for LRLs because it is a sentence-level metric that is more robust to the lack of reference translations than corpus-level metrics like BLEU.  
Although the size has been found to impact model loss rather than performance, \newcite{ghorbani2021scaling} has demonstrated a negative linear relationship between performance and model loss.

\paragraph{Languages} We covered five South Asian languages that are all considered low-resource other than Hindi (\textsc{hi}) \citep{joshi-etal-2020-state}, (Table~\ref{lang-list})\footnote{The classification in \newcite{joshi-etal-2020-state} is outdated. (\textsc{si}) must be at least Joshi's class 3 because it is used to train mBART. According to their definitions, all the languages in our study fall are at least class 2.}; Sinhala (\textsc{si}) and Tamil \textsc{ta} are the official languages of Sri Lanka and Hindi (\textsc{si}), Gujarati (\textsc{gu}), and Kannada (\textsc{ka}) are three of the many official languages of India.  Kannada (\textsc{ka}) is unseen during mBART's pre-training. Note that we only considered the \textsc{en}-XX direction because it often performs better than the XX-\textsc{en} direction \citep{johnson2017googles, lee-etal-2022-pre}. This mitigates our regression models from skewing excessively toward the low spBLEU extreme.
\begin{table*}[!htp]
\centering
\resizebox{\textwidth}{!}{%
\begin{tabular}{lllllllllll}
\hline 
Language & Family & Script & Joshi Class & mBART Token & $d_{geo}$ & $d_{gen}$ & $d_{syn}$ & $d_{pho}$ & $d_{inv}$ & $d_{fea}$ \\
\hline
Kannada (\textsc{ka}) & Dravidian & Kannada & 1 & - & 0.40 & 1.00 & 0.64 & 0.35 & 0.47 & 0.50 \\
Gujarati (\textsc{gu}) & Indo Aryan & Gujarati & 1 & 140M & 0.30 & 0.90 & 0.68 & 0.57 & 0.48 & 0.60 \\
Hindi (\textsc{hi}) & Indo Aryan & Devanagari & 4 & 1715M & 0.40 & 0.90 & 0.59 & 0.34 & 0.47 & 0.50 \\
Sinhala (\textsc{si}) & Indo Aryan & Sinhala & 1 & 243M & 0.40 & 0.90 & 0.78 & 0.41 & 0.50 & 0.60 \\
Tamil (\textsc{ta}) & Dravidian & Tamil & 3 & 595M & 0.40 & 1.00 & 0.71 & 0.57 & 0.50 & 0.60 \\
\hline
\end{tabular}
}
\caption{\label{lang-list} Properties about the languages in our study and their lang2vec distances from English.}
\end{table*}

\paragraph{Corpora} We had two fine-tuning corpora for each language. The first fine-tuning corpus is either an administrative (\textit{Government}; \textsc{si,ta}) or a news (PMIndia; \textsc{hi, gu, ka}) corpus. The second fine-tuning corpus is a religious (\textit{Bible}) corpus. Due to limited availability, we scrapped the Bible corpus for $\textsc{si}$ from a different website\footnote{Sinhala: \url{https://www.wordproject.org/bibles/si/index.htm};
and others: \url{https://ebible.org/download.php}}. For testing corpora, on top of the administrative/ news corpus and religious corpus, we also had an open-domain corpus (\textsc{Flores}). Also due to limited availability, we used a slightly different corpus, $\textsc{Flores-v1}$ instead of $\textsc{Flores-101}$ for $\textsc{si}$. For complete details of the corpora, see Appendix~\ref{dataset_desc}). We define the experiments where the fine-tuning and testing corpora are from the same domain as \textit{in-domain} experiments, and \textit{out-domain} otherwise. 
To ensure that MT systems perform consistently across corpora of varying sizes, we extracted fixed-size fine-tuning sets from each corpus as in Table \ref{raw-expr-records}, based on the available amount of parallel text that we could sample from. All testing corpora are about 1k tokens. 


\paragraph{Data Partitioning} In our modeling, we split our data by grouping them according to their experimental settings (fine-tuning corpus, testing corpus, target language). We refer these groups of experiments as \textit{partitions}. For instance, the "\textsc{ka} partition" refers to the first three columns in Table~\ref{raw-expr-records}, while the "Fined-tuned-on-Bible partition" refers to the last three rows in Table~\ref{raw-expr-records}. We refer the ways of partitioning the data as \textit{partitioning schemes}, which differs by the factor that we model, as in Table~\ref{rmse-results}.\footnote{Partitions with less than 10 data points are too small and thus not discussed.} 


\section{Factors and Featurization of Factors}
We consider three potential factors that impact the performance score of the MT models: 1) the size of fine-tuning corpus, 2) the domain similarity between fine-tuning and testing corpora, and 3) the language similarity between source and target language. We represent these factors as feature variable(s) used as predictor(s) in the regression models described in the next section. 
These predictors are: $\phi_s=$ size feature variable; $\phi_d=$ domain feature variable; $\phi_l=$ language feature variable.

\subsection{Fine-Tuning Corpus Size} It has been observed that the cross-entropy loss of MT models behaves as a power-law with respect to the amount of fine-tuning data \citep{gordon-etal-2021-data, ghorbani2021scaling, kaplan2020scaling}. This suggests that the size of fine-tuning corpora is an important factor to consider in our study. 
We define the size factor, denoted as $\phi_s = \tilde s$, as the normalized count of sentence pairs in the fine-tuning corpus. We achieve this normalization by employing a minimum-maximum scaling method, which constrains it to a range of $0 \leq \tilde s \leq 1$. This standardization aligns with the normalization applied to other features in our study.

\subsection{Domain similarity}
It has been discovered that the performance of language models faces significant drops when they encounter unfamiliar vocabulary and writing style \citep{blitzer2008domain, jia2017adversarial, calapodescu2019sentiment, elsahar-galle-2019-annotate}. We refer to this situation as \textit{domain shift} where \textit{domain} is a "distribution over language characterizing a given topic or genre" \citep{gururangan-etal-2020-dont}. In our case, domain shift happens when the testing corpus is from a domain different from the fine-tuning corpus. This motivates us to consider domain similarity between fine-tuning and testing corpora as one factor affecting the performance of MT models. 

Previous studies have proposed various methods to measure and mitigate domain divergence in MT models \citep{kashyap2021domain, pillutla2021mauve, nayak2023leveraging, lee-etal-2022-pre}. \newcite{kashyap2021domain} showed that information-theoretic measures such as Kullback–Leibler (KL) divergence, Jensen–Shannon divergence (JSD), and higher-order domain discriminator (e.g., Proxy A-distance (\textsc{pad})) capture good correlation with performance drop of MT models. Our study favors entropy methods, particularly JSD over KL divergence and \textsc{pad}, for its symmetric property and relative simplicity. We refer to the domain feature, $\phi_d$, as the JSD between fine-tuning and testing corpora, that is, $\phi_d = j = JSD(t, \tau)$. (see Appendix~\ref{jsd} for complete details on JSD calculation).

\subsection{Language similarity}
Language similarity between source and target languages is important in translating from one language to another because it can help to leverage the cross-lingual transfer and multilinguality of the language model while exploiting parallel data from related language pairs \citep{lee2022improving, gaschi2023exploring, philippy2023towards}. This can be particularly promising for LRLs with insufficient quantities of high-quality parallel data \citep{goyal-etal-2020-efficient}.

To measure language similarity, we utilize six distance features queried from URIEL Typological Database using lang2vec \citep{littell-etal-2017-uriel}. The distance features are geographical distance, $d_{geo}$, genetic distance, $d_{gen}$, syntatic distance, $d_{syn}$, phonological distance, $d_{pho}$, inventory distance, $d_{inv}$, and featural distance, $d_{fea}$ (Table~\ref{lang-list}, see Appendix~\ref{uriel} for details). In our study, we refer to the language feature, $\phi_l$, as any combination of the six distance features.

\section{Methodology}
In this section, we outline our methodology for modeling and evaluating spBLEU predictions using factors mentioned previously, including the exploration of different regression models and their statistical reliability. We also examine the importance of individual features through correlation and feature importance analyses. 

\subsection{Modeling and Evaluation} 
\label{ssec:modeling-and-eval}

Each model is defined by a predictor function $f$, which predicts a spBLEU value given a feature value $x$ or a vector of feature values $\textbf{x}=[x_1, ..., x_n]^\top$ of an experiment. 
Table~\ref{pred_func_tab} catalogues the predictor functions employed. Our selection includes straightforward mathematical functions such as linear, polynomial, and logarithmic types. This choice is grounded in the exploratory nature of our research and the classic use of these functions in regression analysis. It is important to note that in polynomial regressions, interaction variables (for instance, $x_ix_j, i \neq j$) are omitted in multifactor models. This exclusion is deliberate, as it allows us to focus on the impact of individual factors.  The intricate interdependencies among these factors are comprehensively addressed through weight analysis (see \S~\ref{ssec:feat_imp}) in the multifactor linear regression model.

\begin{table}[!htp]
\centering
\resizebox{\columnwidth}{!}{%
\begin{tabular}{ll}
\hline 
Name & Definition \\
\hline
Linear & $f_{\text{lnr}}(\textbf{x}) = \beta_0 + \sum_j \beta_jx_j$ \\
Quadratic &$ f_{\text{poly}_2}(\textbf{x}) = \beta_0 + \sum_j \left [ \beta_{1j} x_j +\beta_{2j} x_j^2 \right] $ \\
Cubic &$ f_{\text{poly}_3}(\textbf{x}) = \beta_0 + \sum_j \left [ \beta_{1j} x_j +\beta_{2j} x_j^2 + \beta_{3j} x_j^3 \right] $ \\
Logarithmic &$ f_{\text{log}}(\textbf{x}) = \beta_0 + \sum_j \beta_j \log x_j$ \\
Scaling Law  & $f_{\text{SL}}(\tilde s) = \beta_0 (\tilde s^{-1} + \beta_1)^{\beta_2}$ \small{(only used for size)}\\
\hline
\end{tabular}
}
\caption{\label{pred_func_tab} The predictor functions explored in our study. 
}
\end{table}

In order to understand the impact of individual factors, we explored predictor functions with one factor at a time as an input variable\footnote{Specifically for size, scaling law was used as an additional predictor function as scaling law as supported by multiple studies \citep{gordon-etal-2021-data, ghorbani2021scaling, kaplan2020scaling}.}.
In addition, data partitioning mentioned in \S~\ref{sec:model-and-data} allowed us to minimize differences between experiments, except for the modeled factor. This approach provides insights into the relationships between individual factors and experimental settings.
 
For further exploration, the same predictor functions were explored using multiple features as multi-factor input variables. This approach allows for a more robust predictor function that captures the interactions between multiple factors, which had been postulated from the partitioning in single-factor modeling. The investigated multi-factor combinations included size and JSD, all six language features, and size, JSD, and all six language features.

To evaluate the prediction accuracy of our regression models, we used root-mean-square error (RMSE) as a metric for ranking models. The RMSE was determined by averaging the RMSE values obtained from each partition's $k$-fold cross-validation folds ($k=10$).

\subsection{Statistical Assessment on Regression Residuals}

Residuals reflect the discrepancy between our model's predicted spBLEU and the true spBLEU for any given experiment. Residuals can provide a quantitative measure of our model's accuracy and how our model's predictions deviate from the true spBLEU, offering insights on any issues with the model's robustness and overall reliability.
We verified two model assumptions described in \newcite{bates-nonlinearregression-1988}, namely, normality and homoscedasticity of residuals.
The normality of residuals is verified using D'Agnostino-Pearson test \cite{pearson}, whereas the homoscedasticity is observed from the plots.


\subsection{Ranking Feature Importance}
\label{ssec:feat_imp}
To assess the correlation between each feature and spBLEU as well as their importance as predictors in our regression models, we ranked the features by the following three analyses:
    
\paragraph{(I) Pearson's Correlation Analysis}
To measure the strength and direction of the linear relationship between each feature and spBLEU, we calculated the Pearson Correlation coefficient along with the statistical significance $p$-value for the correlation. 

\paragraph{(II) Weight analysis}
In addition to pairwise relationships measured by Pearson's Correlation Analysis, we also analyzed the unique contribution of each feature while considering the interdependencies among them by ranking the features by their weight in the multifactor linear regression model.
    
\paragraph{(III) Random Forest}
To assess the importance of each factor in our modeling using various regression models, we used Random Forest to identify the most important features in the multifactor models. See Appendix~\ref{asec:feat-importance-hyperparam} for optimal hyperparameters settings used in our study.

\section{Results}
In this section, we discuss the performance of our regression models based on their RMSE in $k$-fold cross-validation (Table~\ref{rmse-results}). In \S~\ref{ssec:pred-acc}, we extensively discuss the regression models that work well, along with their statistical reliability. Then, in \S~\ref{ssec:residuals-by-partition}, we analyze the residuals' distribution of those models on specific partitions and provide our domain-specific and language-specific interpretations of the observations. Lastly, in \S~\ref{ssec:feat-ranking}, we compare the correlation between each feature and spBLEU, as well as their importance in multifactor models, which gives us insights into the impact of various factors on the performance of MT models.

\subsection{Prediction Accuracy of Factors}
To explore the impact of each factor on spBLEU, we performed regression based on subsets of factors. The prediction accuracy of each regression model was measured in RMSE from $k$-fold cross-validation.

\begin{figure}[ht]
     \centering
     \begin{subfigure}[b]{0.45\textwidth}
         \centering
         \includegraphics[width=\textwidth]{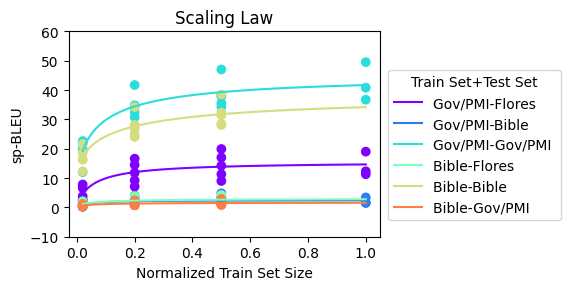}
         \caption{Regression plot using scaling law on size, $f_{\text{SL}}(\tilde s)$; partitioned by both fine-tuning and testing corpora.}
         \label{size-train+test-scaling}
     \end{subfigure}
     \hfill
     \begin{subfigure}[b]{0.45\textwidth}
         \centering
         \includegraphics[width=\textwidth]{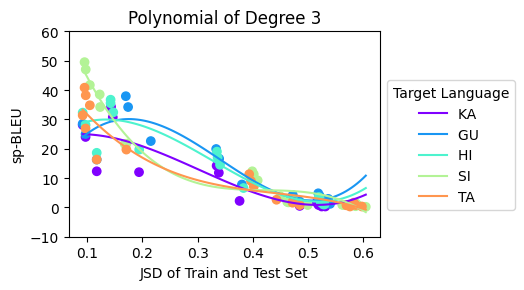}
         \caption{Regression using polynomial (deg 3) regression on JSD, $ f_{\text{poly}_3}(j)$; partitioned by language.}
         \label{jsd-lang-poly3}
     \end{subfigure}
     \hfill
     \begin{subfigure}[b]{0.45\textwidth}
         \centering
         \includegraphics[width=\textwidth]{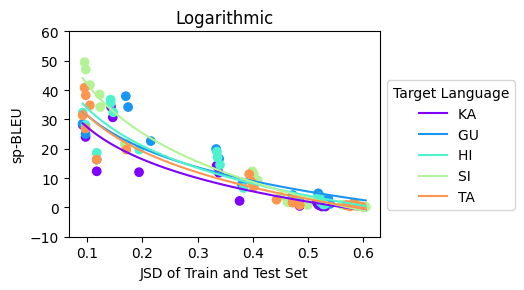}
         \caption{Regression plot using logarithmic regression on JSD, $ f_{\text{log}}(j)$; partitioned by language.}         
         \label{jsd-lang-log}
     \end{subfigure}
        \caption{Regression plots using best predictor functions for size and domain on best partitioning schemes.}
        \label{best-pred-reg-plot}
\end{figure}

\label{ssec:pred-acc}
\paragraph{Regression using size feature} 
In the case of predictor functions that take the size feature as a predictor, we observed that the partitioning scheme has a more significant impact on the RMSE than the predictor functions. For instance, the RMSE is significantly lower when partitioning by fine-tuning and testing corpora (Table~\ref{rmse-results}). Such a trend could be attributed to the concentration of data points when mBART is tested in-domain and out-domain (Figure~\ref{size-train+test-scaling}). Consequently, separating the in-domain and out-domain experiments (i.e., partitioning by both fine-tuning and testing corpora) results in a notably lower RMSE. On the best partitioning scheme, the scaling law model has the lowest RMSE (Figure~\ref{size-train+test-scaling}, RMSE = 2.2998). This result is consistent with the current literature, which asserts that encoder-decoder Transformers used for MT exhibit a scaling law relationship between the volume of training data and model performance.  \citep{gordon-etal-2021-data, ghorbani2021scaling, kaplan2020scaling}.

\begin{table*}[ht]
\centering
\resizebox{\textwidth}{!}{%
\begin{tabular}{l|lllll|ll|l|l}
\hline
\multirow{3}{1in}{Predictor Function} & \multicolumn{9}{c}{Feature Variable(s)$^*$ and partitioning scheme} \\
\cline{2-10}
& \multicolumn{5}{c|}{$\phi_s$ only} & \multicolumn{2}{c|}{$\phi_d$ only} & $\phi_s, \phi_d$ & $\phi_s, \phi_d, \phi_l$\\
 & None & Fine-tune & Test & Lang & Fine-tune, test & None & Lang & None & None\\
\hline
Linear & 13.2388 & 12.9270 & 11.1404 & 13.0014 & \underline{2.9682} & 5.6433 & \underline{5.0782} & 4.8766 & 4.5786\\
Polynomial-2 & 13.2092 & 12.8183 & 11.1218 & 13.0414 & \underline{2.4561} & 5.4633 & \underline{4.5698} & 4.6604 & 4.3840\\
Polynomial-3 & 13.1706 & 12.7914 & 22.4824 & 13.0601 & \underline{2.3335} & \textbf{5.4141} & \underline{\textbf{4.1202}} & \textbf{4.4509} & \textbf{4.2168}\\
Logarithmic & 13.1543 & 12.7835 & 11.3084 & \textbf{12.8578} & \underline{2.3077} & 5.6315 & \underline{4.9247} & 4.9502 & 4.6815\\
Scaling Law & 13.1541 & \textbf{12.7828} & 11.1960 & 12.8929 & \underline{\textbf{2.2998}} & NA & NA & NA & NA\\
\hline 
\end{tabular}
}
\caption{\label{rmse-results} Average Error Measurement$^\dagger$ for Various Prediction Methods and Schemes. \\ 
\footnotesize{* Feature variable(s) used as predictor(s) in the regression models: $\phi_s=$ size feature variable; $\phi_d=$ domain feature variable; $\phi_l=$ language feature variable. \\
$^\dagger$ Measured by average RMSE from $k$-fold cross validation: \textbf{Bold} = function with lowest RMSE on this combination of feature variable(s) and partitioning scheme; \underline{underline} = partitioning scheme with lowest RMSE using this combination of feature variable(s) and predictor function.} 
}
\end{table*}


When modeling with scaling law, the residuals follow normal distribution on all partitions, as in Table~\ref{SL-stats-train+test}. However, the model is heteroscedastic
for partitions involving the Bible corpus that are out-domain. This suggests that translation involving out-of-domain data (particularly Bible corpus) may exhibit highly variable performance. Consequently, it implies that the Bible corpus is better suited for the in-domain corpora rather than out-domain corpora. 

\paragraph{Regression using domain similarity} For predictor functions that take JSD as the predictor, polynomial regression with degree 3 has the lowest RMSE (Figure~\ref{jsd-lang-poly3}, RMSE = 4.1202). Since polynomial regression models have a higher chance of being overfitted as their degree increases, we also consider the best performing non-polynomial model using JSD, i.e., the logarithmic regression model (Figure~\ref{jsd-lang-log}, RMSE = 4.9355). Regarding their statistical reliability, the polynomial regression with degree 3 failed normality test on \textsc{hi} partition while the logarithmic regression failed normality test on \textsc{ta} partition, suggesting specific transformation per language on JSD is needed, otherwise more datapoints is required for the above to ensure model reliability.  

We also noticed that models with size as the predictor have higher RMSE than those with JSD as the predictor. This difference can be attributed to the fact that there are only four unique size values\footnote{For future work, we are collecting more sample points using low-cost transformers.}. Unless we have small enough partitions that contain fewer data points for a fixed size value, for instance, in the fine-tuning-test partition, size as a factor will obtain a lower RMSE.

We also observed that partitioning by language does not lead to a significant improvement in RMSE of the models on either size or JSD. 
This indicates that there is no substantial difference in spBLEU when mBART is tested on various languages, which can be attributed to the limited diversity in our languages. Furthermore, this may suggest a weak correlation between language features and spBLEU as described in Table~\ref{feature-importance}.

\begin{table*}[htp!]
    \begin{subtable}[h]{0.45\textwidth}
        \centering
        \footnotesize{
        \begin{tabular}{l|ll}
        \hline
        Fine-tuning -- test & Normality & Homoscedastic? \\
        \hline
        bible-bible & 0.3996 & Yes \\
        bible-\textsc{flores} & 0.1380 & \textbf{No} \\
        bible-gov & 0.2570 & \textbf{No} \\
        gov-bible & 0.2534 & \textbf{No} \\
        gov-\textsc{flores} & 0.2623 & Yes \\
        gov-gov & 0.6127 & No \\
        \hline
        \end{tabular}
        }
       \caption{$f_{\text{SL}(\tilde s)}$ on each train-test partition.}
       \label{SL-stats-train+test}
    \end{subtable}
    \quad
    \begin{subtable}[h]{0.55\textwidth}
        \centering
        \footnotesize{
        \begin{tabular}{l|ll|ll}
        \hline
        & \multicolumn{2}{|c|}{\scriptsize{$f_{\text{poly3}}(j)$}} & \multicolumn{2}{c}{\scriptsize{$f_{\text{log}}(j)$}} \\
        \hline
        \scriptsize{Language} & \scriptsize{Normality} & \scriptsize{Homoscedastic?} & \scriptsize{Normality} & \scriptsize{Homoscedastic?} \\
        \hline
        \textsc{ka} & 0.1578 & Yes & 0.2155 & Yes \\
        \textsc{gu} & 0.0563 & Yes & 0.2027 & Yes \\
        \textsc{hi} & \textbf{0.0129} & Yes & 0.7290 & Yes \\
        \textsc{si} & 0.6021 & Yes & 0.2702 & Yes \\
        \textsc{ta} & 0.0500 & Yes & \textbf{0.0299} & Yes \\
        \hline
        \end{tabular}
        }
        \caption{$f_{\text{poly3}}(j)$ and $f_{\text{log}}(j)$ for each language partition.}
        \label{JSD-stats}
     \end{subtable}
     \caption{Statistical Assessment on Normality and Homoscedasticity for size and JSD on best partitioning schmes respectively.
     \footnotesize{For normality, \textbf{bold} = residuals are not normally distributed $(p<0.05)$. 
     }
     }
     \label{impprtant-stats}
\end{table*}

\paragraph{Regression using multiple factors} We evaluated two additional regression models with multiple factors to examine how these factors interact with each other in predicting spBLEU scores. Table~\ref{rmse-results} includes RMSE of multifactor models with $\phi_s$ and $\phi_d$ as predictors, and multifactor models with $\phi_s$, $\phi_d$, and $\phi_l$ (all lang2vec distances in Table~\ref{lang-list}) as predictors.

Relative to single-factor models that take only $\phi_d$ without partitioning, we observed that including $\phi_s$ and $\phi_l$ does improve the RMSE. However, the improvement is insignificant, further suggesting the high importance of domain similarity in the prediction relative to other factors considered in this study.

\subsection{Residuals by Partition}
\label{ssec:residuals-by-partition}

\begin{figure*}[!ht]
    \centering
    \begin{subfigure}[b]{0.3\textwidth}
        \centering
        \includegraphics[width=\textwidth]{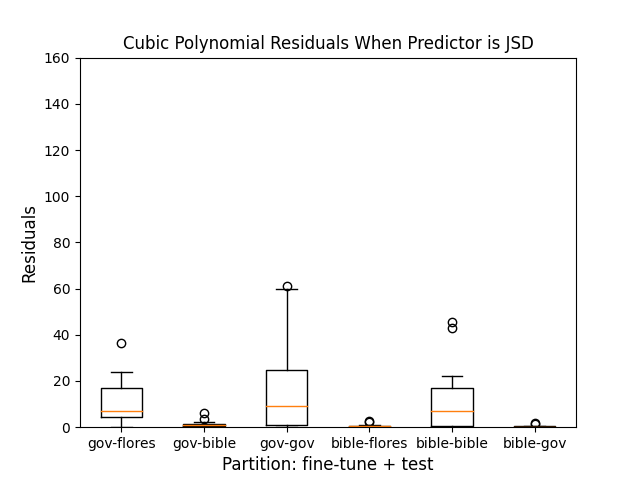}
        \caption{Residuals from $f_{\text{SL}}(\tilde s)$; partitioned by fine-tuning and testing corpora.}
        \label{boxplot-size-traintest-scaling}
    \end{subfigure}
    \quad
    \begin{subfigure}[b]{0.3\textwidth}
        \centering
        \includegraphics[width=\textwidth]{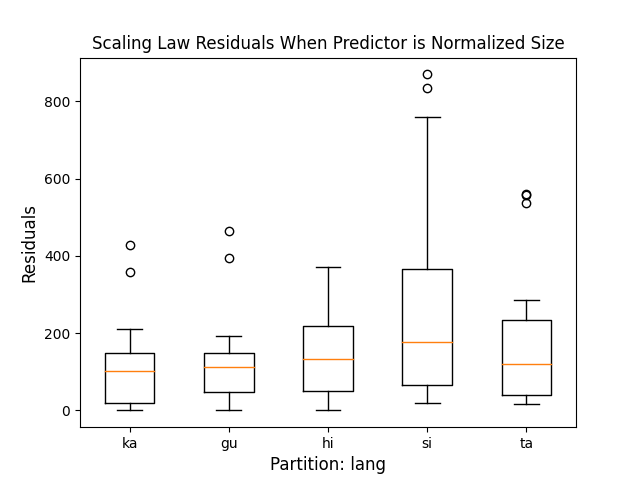}
        \caption{Residuals from $f_{\text{SL}}(\tilde s)$; partitioned by language.}
        \label{boxplot-size-lang-scaling}
    \end{subfigure}
    \quad
    \begin{subfigure}[b]{0.3\textwidth}
        \centering
        \includegraphics[width=\textwidth]{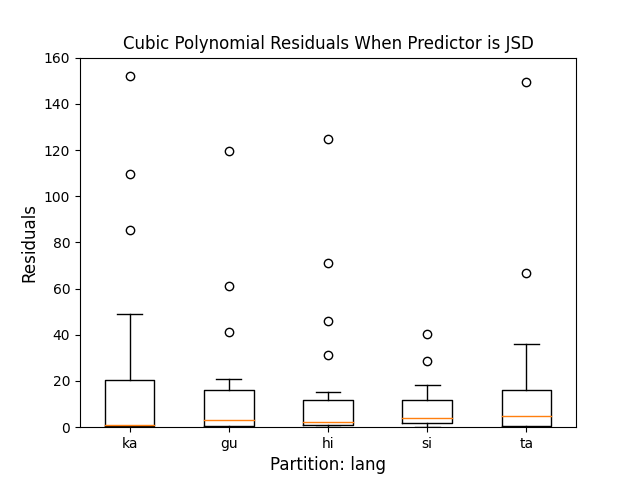}
        \caption{Residuals from $ f_{\text{poly}_3}(j)$; partitioned by language.}
        \label{boxplot-jsd-lang-poly3}
    \end{subfigure}
    \caption{Boxplots of residuals using best predictor functions for size and domain on some partitioning schemes.}
    \label{best-pred-reg-plot}
\end{figure*}

To observe how our models performs on different partitions, we created boxplots of residuals when modeling data on each partition using the predictor functions. Using the best predictor function for size (scaling law) with the best partitioning scheme (by both fine-tuning and testing corpora), we noticed that the mean and variance of the residuals were lower for out-domain partitions (gov-gov and bible-bible, Figure~\ref{boxplot-size-traintest-scaling}). This suggests that our model predicts better for out-domain partitions, which could be explained by the difference in the range of raw spBLEU when mBART is tested on in-domain and out-domain experiments ([$6.5, 49.5$] for in-domain, [$0.2, 19.9$] for out-domain).

Figure~\ref{boxplot-size-lang-scaling} presents how well the scaling law works for different languages. We noticed that the $\textsc{si}$ partition has relatively high residual mean and variance, implying that the performance of mBART on Sinhala is harder to predict with respect to the size of the fine-tuning corpus. This could be due to the use of different versions of the Bible corpus and $\textsc{Flores}$ corpus for $\textsc{si}$, resulting in a higher range of spBLEU in this partition ([0.2, 49.5], Table~\ref{raw-expr-records}) and hence harder to predict. However, this phenomenon is not observed in Figure~\ref{boxplot-jsd-lang-poly3} when the feature variable is JSD. This implies that using JSD as the predictor yields a more stable prediction for \textsc{si} because it is not affected by using different fine-tuning corpora.

\subsection{Feature Rankings}
\label{ssec:feat-ranking}

In order to assess the impact of the features in predicting spBLEU, Table~\ref{feature-importance} provided Pearson correlation coefficient and the statistical significance measured in $p$-value. We also include weights for each feature in the best multifactor linear regression model computed and their feature importance based on the best-performing Random Forest Regressor.

In Pearson's Correlation Analysis ranking, JSD stands out with a strong and statistically significant correlation to spBLEU (Table~\ref{feature-importance}), suggesting a strong linear relationship between JSD and spBLEU. It also ranks highest in both weight analysis and Random Forest feature importance analysis, further illustrating its importance in predicting spBLEU (Table~\ref{feature-importance}). This finding brings hope for developing a reliable model to understand the relationship between domain similarity and performance in MT tasks. 

Surprisingly, all six language features show low correlations with spBLEU. The high similarity amongst our South Asian languages could be a factor, resulting in a similar distance from \textsc{en} in Table~\ref{lang-list}. It suggests that the language features are not as significant as other features, like size and domain, for use as predictors in our regression models.

\begin{table}[!htp]
\centering
\fontsize{9pt}{11pt}\selectfont
\resizebox{\columnwidth}{!}{%
\begin{tabular}{m{0.1\columnwidth}|m{0.2\columnwidth}m{0.25\columnwidth}|m{0.25\columnwidth}|m{0.2\columnwidth}}
\hline
Feature Variable & Pearson Correlation Coefficient & Statistical Significance ($p$-value) & Weight Analysis & Random Forest (\%) \\
\hline
$j$ & -0.9176 [1] & $8.47 \times 10^{-71}$ & -68.5404 [1] & 88.393 [1] \\
$\tilde s$ & 0.2468 [2] & 0.0010& 19.1317 [3=] & 7.805 [2] \\
$d_{gen}$ & -0.0863 [3] & 0.2574 & -25.7118 [2] & 0.365 [5] \\
$d_{syn}$ & 0.0365 [4] & 0.6325 & 3.6204 [7] & 2.267 [3] \\
$d_{inv}$ & 0.0239 [6] & 0.7542 & 13.0297 [5] & 0.782 [4] \\
$d_{fea}$ & 0.0337 [5] & 0.6585 & 19.1317 [3=] & 0.079 [8] \\
$d_{geo}$ & 0.0025 [7] & 0.9738 & 7.1308 [6] & 0.147 [7] \\
$d_{pho}$ & -0.0076 [8] & 0.9104 & -1.1780 [8] & 0.161 [6] \\
\hline
\end{tabular}
}
\caption{\label{feature-importance} Feature importance rankings by Pearson's correlation analysis (along with its statistical significance), weight in linear regression model, and Random Forest feature importance analysis. Rankings in brackets.}
\end{table}

\section{Discussion}
In this study, we revealed that domain similarity plays an important role in MT. In other words, it significantly affects the performance of MT models. All three feature rankings in \S~\ref{ssec:feat-ranking}, as depicted in Table~\ref{feature-importance}, underscore the significance of domain similarity in predicting spBLEU. The relationship between JSD and spBLEU is best modeled by polynomial regression of degree 3 in terms of $k$-fold RMSE, whereas the best non-polynomial model was logarithmic regression. Both models are relatively reliable in terms of the normality and homoscedasticity of the residuals.

Recognizing the importance of domain similarity in MT, we also observed how it affects the predictability of spBLEU when modeling with the scaling law, which uses size as a predictor. The separation of in-domain and out-domain data improves the RMSE due to the distinct clustering of in-domain and out-domain data points. Additionally, we found that the performance of MT models on out-domain partitions is easier to predict. In other words, the prediction models are more confident that the spBLEU values are low when the range of spBLEU values is small. However, despite the lower variance in the residuals of the scaling law on out-domain partitions, the residuals exhibit heteroscedasticity in most of the out-domain partitions when using the scaling law for modeling. 

Furthermore, the \textsc{Flores}-v1 dataset for Sinhala includes data from OpenSubtitles, which are mainly transcripts of spoken data (\newcite{guzman-etal-2019-flores, lison2018opensubtitles2018}). It should be noted that these transcripts may exhibit varying degrees of reliability, as they lack a control mechanism for verifying the translation accuracy. In addition, spoken Sinhala has different syntactical rules of written Sinhala \cite{de2019survey}), which means that there is variation in our Sinhala corpus (e.g., Bible and government documents corpora) as well. This would likely result in a lower translation score across FLORES-v1 and out-domain corpus. However, the JSD score can predict some of these differences in language caused by domain shift, similar to partitioning out by fine-tuning and test datasets. This explains why our model’s predictive performance improved under these conditions. 

Additionally, the Sri Lanka constitution states that “Sinhala shall be the language of administration and be used for the maintenance of public records and the transaction of all business” for most regions (\newcite{constsrilanka2923}). Tamil, also an official language of Sri Lanka, would instead be translated. This difference in language choice could also explain why Sinhala outperforms Tamil in government-related in-domain documents and why domain similarity is such a powerful predictor in these cases. 

Furthermore, we have detected heteroscedasticity in various models. For JSD, the data points will be heteroscedastic due to the inherent high domain divergence, resulting in experiments with very low spBLEU. In contrast, low domain divergence is highly variable, as other factors, such as language and fine-tuning set size, can impact the MT performance. The observation that JSD does not guarantee good model performance in single-factor regression motivates us to consider alternative techniques. The alternative techniques should be more robust or include additional variables to capture variations during low-JSD predictions. Additionally, we observed from the boxplots of residuals that residuals are skewed towards low spBLEU. 

\section{Conclusion}
In our research, we conducted a comprehensive analysis focusing on three key factors (the size of the fine-tuning corpus, domain similarity between the fine-tuning and testing corpora, and the linguistic similarity between the source and target languages) affecting performance prediction of the MT for five South Asian languages. We find that domain similarity exerts the most significant influence on performance, surpassing even the impact of fine-tuning the corpus size. Additionally, the background of the corpora and language being translated emerged as a crucial factor in predicting performance and stability. Lastly, we verify that our approach to ascertain predictive factors for LRLs' performance is statistically rigorous. This approach enables performance prediction without the need for fine-tuning and testing resource-intensive and costly language models, ultimately fostering greater accessibility and equity for LRLs.

\subsection*{Limitations}
The most prominent limitation of our study is the amount of data to fine-tune our regression models. As we observed that our models are generally biased towards experiments with low spBLEU and we could include more experiments with larger fine-tuning corpus size, or perhaps at constant interval between 1k and 100k tokens. There could also be a need to balance the amount of data from in-domain and out-domain.  

The high degree of similarity between the languages in our data set rendered the effectiveness of language features from lang2vec as predictors. Due to the lack of LRL data in the URIEL library, lang2vec may not have sufficient data to provide approximation that accurately describe the LRL. Consequently, many languages might exhibit similar values for the same features, making it difficult to distinguish between them. This motivates us to consider incorporating experiments involving a more diverse range of languages in future studies in order to thoroughly examine the impact of language similarity on MT. Additionally, apart from dataset-independent linguistic features, as suggested by \newcite{lin2019choosing}, we will explore dataset-dependent language features (e.g., Type-Token Ratio (TTR), word overlap, and subword overlap). Therefore, a more rigorous investigation into measuring language similarity is essential to identify suitable predictors for our task.

In addition, it is also important to consider additional factors that could potentially impact the performance of MT models, such as the use of pivot languages \citep{srinivasan2021predicting} and the presence of noise \citep{gordon-etal-2021-data}. Expanding our analysis to include data from different MT models and various evaluation metrics will help us assess the transferability of our prediction models across different MT models and evaluation metrics.

\subsection*{Acknowledgement}
We extend our profound gratitude to the Fields Undergraduate Summer Research Program (FUSRP) for their invaluable support and the unique opportunity they provided for engaging in high-quality mathematical research. Our sincere thanks also go to Juan Armando Parra Flores and Leandro Arcos Roman, whose contributions through the FUSRP were instrumental in the success of our work. 

\subsection*{Ethical Considerations}
\paragraph{Equitability in Language Representation} Given that our study revolves around LRLs, it is imperative to conscientiously acknowledge the imperative to foster equitable technological developments across varied linguistic communities. Our exploration into optimizing MT models for LRLs partially addresses this, but it’s vital to consistently prioritize and amplify underrepresented languages in our future research and model development to prevent linguistic bias and facilitate digital inclusivity.

\paragraph{Data Bias and Representation}Our regression models, as indicated in the limitations section, have potential biases towards experiments with low spBLEU, which may affect the robustness and fairness of our predictive models across various language datasets and use-cases. Ensuring unbiased and representative datasets is crucial not only for the accuracy of predictive models but also for avoiding the unintentional marginalization of certain linguistic features or dialects within the LRLs.

\bibliographystyle{acl_natbib}
\bibliography{eacl}

\appendix

\section{Experimental Setup}

\subsection{Details of Corpora}
\label{dataset_desc}

\subsection*{Bible corpus (Bible)} The JHU Bible Corpus \citep{mccarthy-etal-2020-johns} contains Bible translations in over 1600 languages and serves as the only available parallel text for several low-resource languages. Due to the limited data available for our languages, we created a Bible corpus specifically for our experiments by scrapping Bible data from web\footnote{Sinhala: \url{https://www.wordproject.org/bibles/si/index.htm};
and others: \url{https://ebible.org/download.php}} and aligned the sentences at verse level automatically. The resulting curated multi-way parallel corpus consists of 25k parallel sentences in \textsc{ka, gu, hi}, and \textsc{ta}. Note that \textsc{si} was sourced from a different website, resulting in distinct content for this language.


\subsection*{\textsc{Flores} corpus}
{\textsc{Flores}-101 (Flores)} \citep{goyal-etal-2022-flores} is a corpus containing translations of English Wikipedia sentences into 101 different languages. The translations were done manually, and the corpus covers diverse topics and domains. For \textsc{si}, we use \textsc{Flores}-v1 \citep{guzman-etal-2019-flores} instead since it is not present in \textsc{Flores}-101 .

\subsection*{Government corpus (Gov)} 
The government corpus (Gov) \citep{fernando2021data} is a multi-way parallel corpus comprising Sinhala, Tamil, and English texts. The corpus is manually curated and includes data from various official Sri Lankan government sources, such as annual reports, committee reports, government institutional websites, procurement documents, and acts of the Parliament.

\subsection*{PMIndia corpus (PMI)} 
The PMIndia corpus (PMI) \citep{haddow2020pmindia} is a multi-way parallel corpus consisting of 13 Indian languages, along with English. The corpus has been curated from news updates taken from the Prime Minister of India's website.

\subsection{Jensen-Shannon Divergence}
\label{jsd}
Jensen-Shannon divergence (JSD) beteen two distributions $P$ and $Q$ is calculated using the formula
$$
    JSD(P||Q) = \frac{1}{2} KL (P||M) + \frac{1}{2} KL(Q||M)
$$
where $M$ is an equally weighted sum of the wo distributions and $KL (\cdot || \cdot)$ is the Kullback-Leibler divergence. 

In preparation of this calculation, we first tokenized each corpus using the NLTK package\footnote{Documentation of NLTK package: \url{https://www.nltk.org/}}, striped all stopwords, and transformed them into a (discrete) frequency distribution over all word tokens. Then, we convert all times and numbers into the tokens \texttt{<TIME>} and \texttt{<NUMBER>}, respectively. Finally, we compared the frequency distributions of each fine-tuning and test set using the formula above.

Note that JSD ranged from 0 to 1, with lower values indicating higher similarity between the two distributions.

\subsection{Language Features}
\label{uriel}
In this study, language feature refers to measures of similarity between two languages that are based on phylogenetic or typological properties established by linguistic study. The six language features from the URIEL database \newcite{littell-etal-2017-uriel} utilized in this study includes:

\subsection*{Geographic distance ($d_{geo}$)} 
The orthodromic distance between the languages on the surface of the earth, divided by the antipodal distance. It is based primarily on language location descriptions in Glottolog \citep{hammarstrom2018glottolog}.

\subsection*{Genetic distance ($d_{gen}$)} 
The genealogical distance of the languages, derived from the hypothesized tree of language descent in Glottolog.

\subsection*{Inventory distance ($d_{inv}$)} 
The cosine distance between the phonological feature vectors derived from the PHOIBLE database \citep{moran2014phoible}.

\subsection*{Syntactic distance ($d_{syn}$)}
The cosine distance between the syntactic structures feature vectors of the languages \citep{chris-syntactic}, derived mostly from the WALS database \citep{wals}.

\subsection*{Phonological distance ($d_{pho}$)}
The cosine distance between the phonological feature vectors derived from the WALS and Ethnologue databases \citep{ethnologue}.

\subsection*{Featural distance ($d_{fea}$)}
The cosine distance between feature vectors combining all 5 features mentioned above.

\section{Hyperparameters of Random Forest Regressor}
\label{asec:feat-importance-hyperparam} 

We conducted grid search with $k$-fold cross-validation to find the optimal hyperparameter settings, including the number of decision trees in the ensemble (\texttt{n\_estimators}), the maximum depth of each decision tree (\texttt{max\_depth}), the minimum number of samples required to split an internal node (\texttt{min\_samples\_split}), the minimum number of samples required to be at a leaf node (\texttt{min\_samples\_leaf}), and whether bootstrap samples were used in building trees (\texttt{bootstrap}). The optimal hyperparameter settings are tabulated in Table~\ref{feat-importance-hyperparam}, resulting in an RMSE of 3.29. 
\begin{table}[ht]
\centering
\resizebox{\columnwidth}{!}{%
\begin{tabular}{lll}
\hline
\textbf{Hyperparameter} & \textbf{Values Searched} & \textbf{Optimal Setting} \\
\hline
\texttt{n\_estimators} & $\{n \,|\, n = 50 + 25k, \, 0 \leq k \leq 14\}$ & 100 \\
\texttt{max\_depth} & $\{n \,|\, n = 3 + 2k, \, 0 \leq k \leq 6\}$ & 9\\
\texttt{min\_samples\_split} & $\{2, 3, 4, 5\}$ & 1 \\
\texttt{min\_samples\_leaf} & $\{1, 2, 3\}$ & 2\\
\texttt{bootstrap} & $\{\textsc{true, false}\}$ & \textsc{true}\\
\hline
\end{tabular}
}
\caption{\label{feat-importance-hyperparam} List of hyperparameters used in the optimization of the Random Forest Regressor using grid search.}
\end{table}

\onecolumn
\newpage 

\section{Scatter Plots}
In this section, we present the scatter plots of spBLEU with respect to size of fine-tuning corpora using different partitioning schemes.
\label{asec:scatter-plots}
\subsection{Factor = Size}

\begin{figure*}[!htp]
    \centering
    \begin{subfigure}[b]{0.45\textwidth}
        \includegraphics[width =\textwidth]{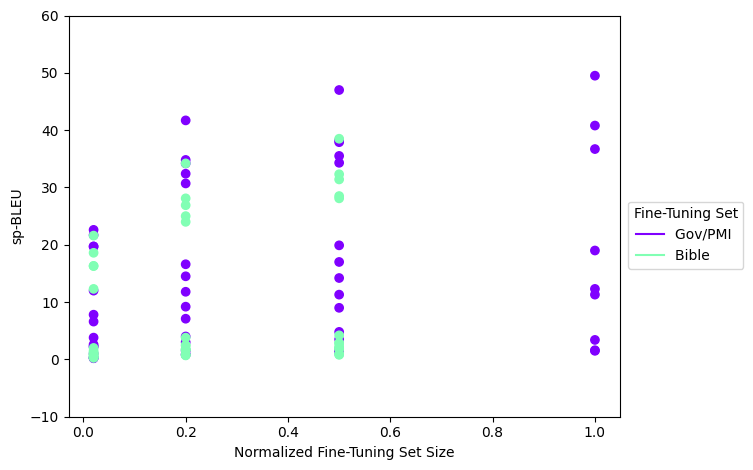}
        \centering
        \caption{\label{scatter-nsize-train}Scatter Plot of spBLEU with respect to size, partitioned by fine-tuning corpora.}
    \end{subfigure}
    \quad
    \begin{subfigure}[b]{0.45\textwidth}
        \includegraphics[width =\textwidth]{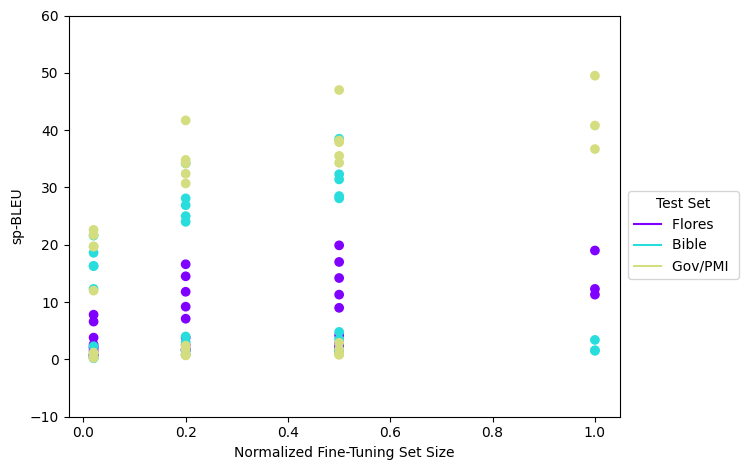}
        \centering
        \caption{\label{scatter-nsize-test}Scatter Plot of spBLEU with respect to size of fine-tuning corpora, partitioned by testing corpora.}
    \end{subfigure}
    \\
    \begin{subfigure}[b]{0.45\textwidth}
        \includegraphics[width =\textwidth]{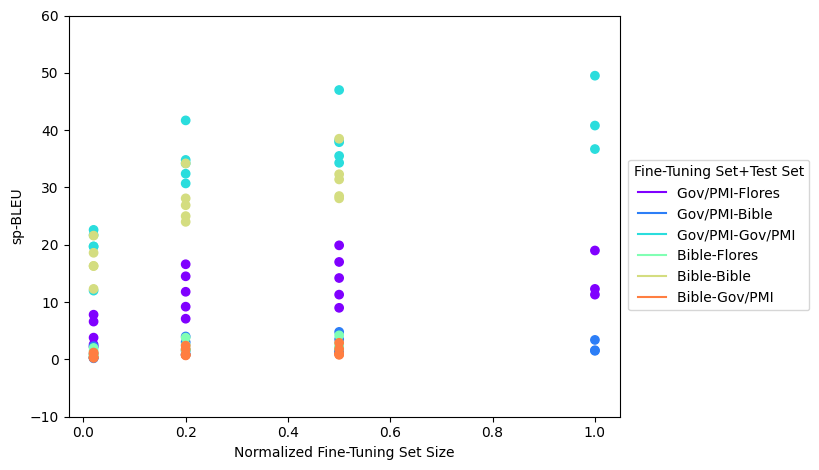}
        \centering
        \caption{\label{scatter-nsize-traintest}Scatter Plot of spBLEU with respect to size, partitioned by both fine-tuning and testing corpora.}
    \end{subfigure}
    \quad
    \begin{subfigure}[b]{0.45\textwidth}
        \includegraphics[width =\textwidth]{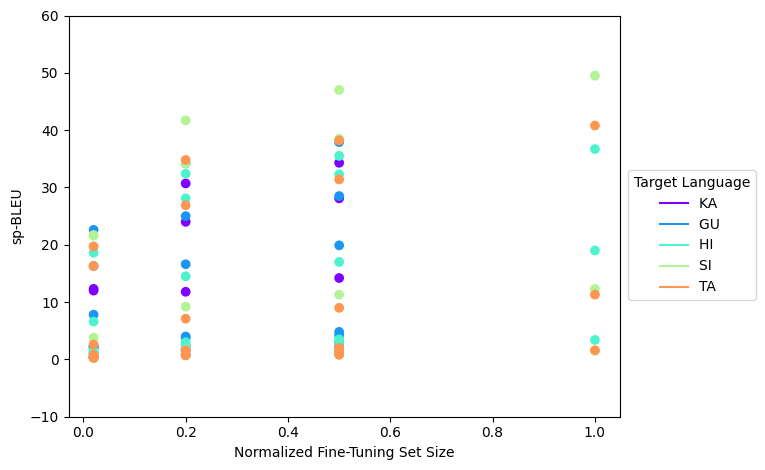}
        \centering
        \caption{\label{scatter-nsize-lang}Scatter Plot of spBLEU with respect to size, partitioned by target language.}
    \end{subfigure}
    \caption{Scatter Plots of spBLEU with respect to size using different partitioning schemes.}
    \label{scatter-nsize-all}
\end{figure*}

\subsection{Factor = Domain Similarity}
In this section, we present the scatter plot of spBLEU with respect to JSD, partitioned by target language.

\begin{figure}[h]
    \includegraphics[width =0.45\columnwidth]{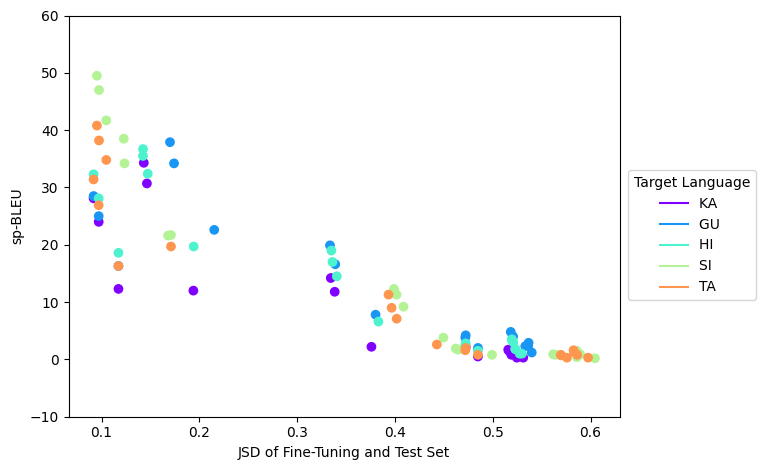}
    \centering
    \caption{\label{scatter-jsd-lang}Scatter Plot of spBLEU with respect to JSD, partitioned by target language.}
\end{figure}

\end{document}